\pdfoutput=1

\documentclass{article}

\usepackage[final]{neurips_2024}

\usepackage{times}
\usepackage{latexsym}

\usepackage[T1]{fontenc}

\usepackage[utf8]{inputenc}

\usepackage{microtype}

\usepackage{inconsolata}

\usepackage{graphicx}
\usepackage{subcaption}
\usepackage{booktabs} 
\usepackage{textcomp}

\usepackage{amsmath}
\usepackage{amssymb}
\usepackage{mathtools}
\usepackage{amsthm}
\usepackage{multirow}
\usepackage{tabularx}
\usepackage{longtable}
\usepackage{wrapfig}

\newcolumntype{b}{X}
\newcolumntype{s}{>{\hsize=.4\hsize}X}
\newcolumntype{d}{>{\hsize=.15\hsize}X}
\usepackage[frozencache,cachedir=.]{minted}
\usepackage{listings}
\usepackage{caption}
\usepackage{adjustbox}
\usepackage{xcolor} 
\definecolor{LightGray}{gray}{0.9}
\usepackage{breqn}
\usepackage{enumitem}
\usepackage{dblfnote}

\definecolor{darkblue}{rgb}{0, 0, 0.5}
\usepackage{hyperref}
\hypersetup{colorlinks=true, citecolor=darkblue, linkcolor=darkblue, urlcolor=darkblue}
\usepackage[capitalize,noabbrev]{cleveref}

\theoremstyle{plain}

\theoremstyle{definition}

\theoremstyle{remark}

\usepackage[textsize=tiny]{todonotes}

\usepackage{algorithm}
\usepackage{algorithmic}

\usepackage{pifont}
\graphicspath{ {./figs/} }
\newenvironment{longlisting}{\captionsetup{type=listing}}{}
\usepackage{breqn}

\title{Learning Metadata-Agnostic Representations for Text-to-SQL In-Context Example Selection}

\author{Chuhong Mai$^*$ \\
  Amazon Web Services \\
  \texttt{maichuh@amazon.com} \\\And
  Ro-ee Tal$^*$ \\
  Amazon Web Services \\
  \texttt{rttal@amazon.com} \\\And
  Thahir Mohamed \\
  Amazon Web Services \\
  \texttt{thahirm@amazon.com} \\}

\begin{document}
\maketitle
\def\thefootnote{*}\footnotetext{These authors contributed equally to this work}\def\thefootnote{\arabic{footnote}}
\begin{abstract}
In-context learning (ICL) is a powerful paradigm where large language models (LLMs) benefit from task demonstrations added to the prompt.
Yet, selecting optimal demonstrations is not trivial, especially for complex or multi-modal tasks where input and output distributions differ.
We hypothesize that forming task-specific representations of the input is key. 
In this paper, we propose a method to align representations of natural language questions and those of SQL queries in a shared embedding space.
Our technique, dubbed MARLO—\textbf{M}etadata-\textbf{A}gnostic \textbf{R}epresentation \textbf{L}earning for Text-t\textbf{O}-SQL— uses query structure to model querying intent without over-indexing on underlying database \textbf{metadata} (i.e. tables, columns, or domain-specific entities of a database referenced in the question or query).
This allows MARLO to select examples that are structurally and semantically relevant for the task rather than examples that are spuriously related to a certain domain or question phrasing.
%
When used to retrieve examples based on question similarity, MARLO shows superior performance compared to generic embedding models (on average +2.9\%pt. in execution accuracy) on the Spider benchmark.
It also outperforms the next best method that masks metadata information by +0.8\%pt. in execution accuracy on average, while imposing a significantly lower inference latency.
\end{abstract}

\begin{figure*}[ht]
    \vspace{-1em}
    \centering
    \includegraphics[width=\linewidth]{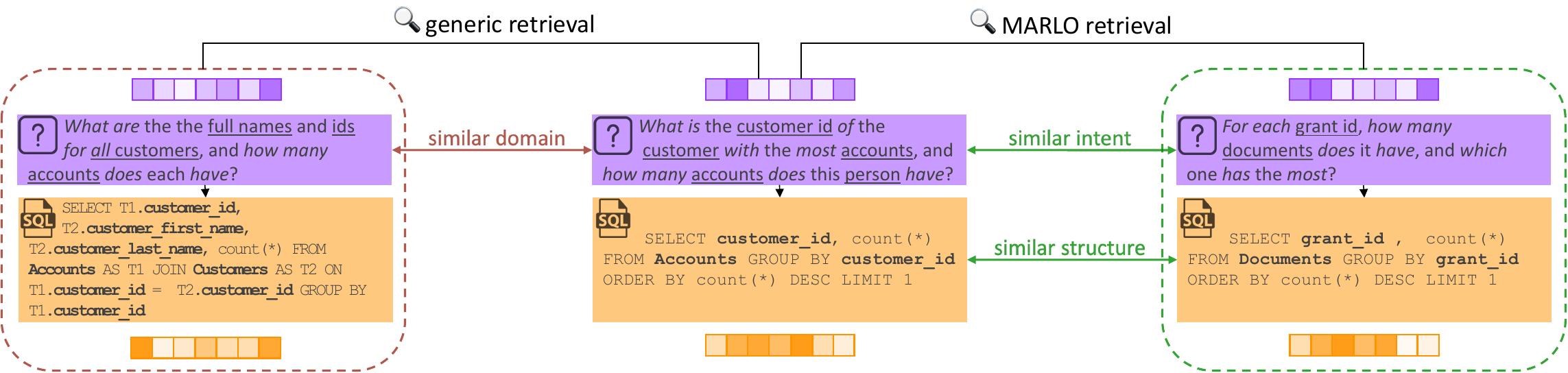}
    \caption{\textbf{Motivation of this work}. 
    From the perspective of generic sentence embeddings, the left question is similar to the middle one but dissimilar from the one on the right. MARLO focuses on query structure (rather than metadata specifics) to represent the intent of each question more accurately. This allows it to retrieve the more instructive demonstration (rightmost). For emphasis, \underline{noun chunks}, \textit{parts-of-speech}, and \textbf{domain information specific to the database metadata} are annotated accordingly.}
    \label{fig:diagram}
    \vspace{-1em}
\end{figure*}

\section{Introduction}
\label{sec:intro}
Large Language Models (LLMs) have demonstrated significant few-shot performance gains on a variety of NLP tasks simply by conditioning on example demonstrations during inference—an approach commonly referred to as in-context learning (ICL) \cite{brown2020language, dong2023survey}.
Not only is this little-understood emergent ability \cite{wei2022emergent} an active area of research \cite{akyürek2023learning, NEURIPS2022_c529dba0, shin-etal-2022-effect, min2022rethinking, xie2022explanation}, further effort has been made to understand why LLM performance remains sensitive to demonstration selection \cite{liu-etal-2022-makes, gao2023texttosql, an-etal-2023-skill, wang2023large}, ordering \cite{lu2022fantastically}, and formatting \cite{chen2023unleashing, gao2023texttosql}.
Inspired by the success of retrieval-augmented generation on knowledge intensive tasks \cite{NEURIPS2018_cd17d3ce, 10.5555/3495724.3496517, karpukhin-etal-2020-dense, gao2023retrievalaugmented}, recent works demonstrate the effectiveness of selecting in-context examples that are semantically similar to the test exemplar \cite{rubin-etal-2022-learning, liu-etal-2022-makes, duan2023exploring}.
However, semantic textual similarity has its challenges in the context of demonstration retrieval, particularity when the task is multi-modal (e.g. text $\mapsto$ image) or its input and output distributions differ (e.g. English $\mapsto$ French or SQL).

In this paper, we specifically focus on demonstration retrieval for Text-to-SQL as it is common practice in SOTA systems, and because the semantic and syntactic representations of natural language questions and structured queries are inherently difficult to compare.
Generic retrievers tend to select examples by matching the domain or nouns in questions that relate to tables, columns, or entities of a database (e.g. `accounts' or `customers' in \autoref{fig:diagram}), which we refer to broadly as database \textbf{metadata}.
As a result, selected demonstrations describe semantically similar entities but have unrelated intent or query structure.
To overcome this limitation, we fine-tune a pre-trained language model to align representations of natural language questions and those of SQL queries in a shared embedding space based on the question intent and corresponding query structure.
Using a novel \textbf{metadata}-agnostic metric we propose (\autoref{eq:qed}), we learn representations that pay closer attention to structural information and retrieve lexically varied yet semantically meaningful demonstrations.

Our work, dubbed \textbf{M}etadata \textbf{A}gnostic \textbf{R}epresentation \textbf{L}earning for Text-t\textbf{O}-SQL (MARLO), differs from prior works that i) compare embeddings derived from heuristic feature extraction \cite{Makiyama2015TextMA, 10.1007/s10115-013-0614-1, 8352666}, ii) use general-purpose retrievers to select from suitably formatted prompts \cite{sun2023enhancing}, iii) use custom retrievers that add special code tokens to their vocabulary \cite{tipirneni2022structcoder} or are fine-tuned on masked task-specific data \cite{gao2023texttosql}, or iv) annotate demonstrations by hand, which is prohibitively expensive and inflexible \cite{pourreza2023dinsql}.


From our ablation analysis and experiments, we observe example selection with MARLO outperforms alternative methods discussed in literature, including the state-of-the-art for the Text-to-SQL task on benchmark dataset.
Compared to multi-stage masking approaches that require multiple LLMs calls, MARLO performs better or on par with lower inference latency.
In summary our major contributions are:
\begin{enumerate}[leftmargin=*]
    \item A novel method to jointly learn aligned embeddings for natural language questions and SQL queries using a query edit distance heuristic. The customized embedding model is able to better comprehend the anticipated SQL structure associated with natural language questions.
    \item An in-depth analysis of ICL example selection methods for Text-to-SQL, highlighting how the choice of selected examples is particularly useful for question understanding in this task and why MARLO works. 
    \item First comprehensive showcase of competitive instruction-tuned foundation models on this task.
\end{enumerate}

\section{Related Work}
\label{background}

\subsection{ICL Demonstrations for Text-to-SQL}
\label{subsec:icl-demos}
Several lines of work have explored what aspects of a demonstration contribute to ICL performance gains --
the input distribution (e.g. formatting, overall perplexity) \cite{min2022rethinking, gonen-etal-2023-demystifying};
the semantic similarity between demonstrations and the inference exemplar \cite{liu-etal-2022-makes, duan2023exploring, qin2023incontext}; and
diversity \cite{qin2023incontext}.
For a skill-intensive task like Text-to-SQL, similarity is a more important dimension than diversity. 
Demonstrations that possess similar questions \cite{liu-etal-2022-makes} or SQL queries \cite{nan2023enhancing} to those of the inference exemplar are more helpful than random selections.
However, solely relying on embeddings of raw input questions or queries for similarity retrieval often suffers from the bias of surface language features (e.g. nouns that contain domain information).
\citealt{guo2023case} and DAIL-SQL \cite{gao2023texttosql} address this by masking noun chunks in the question or query referenced in the metadata before encoding.
Skill-KNN \cite{an-etal-2023-skill} take an alternative approach by rewriting the questions as skill descriptions which, rather than the questions, are encoded and retrieved using generic embedding models.
To the best of our knowledge, we are the first to tackle this problem by learning embeddings for natural language questions and SQL queries that are aligned in a shared space and represent the user's intent agnostic to the domain or any referenced database metadata.

\subsection{Similarity-Based Retrieval}
\label{subsec:retrieving-demos}
Similarity-based retrieval is typically framed as a metric learning problem that uses dot-product or cosine similarity as a ranking function \cite{8186753}.
In cases where input and output distributions are distinct, it is a common practice to fine-tune or align pre-trained transformers using language modeling or constrastive objectives, often in a Siamese configuration
\cite{cer-etal-2018-universal, reimers-gurevych-2019-sentence, yang-etal-2020-multilingual, gao-etal-2021-simcse, ni-etal-2022-sentence, neelakantan2022text}.
For example, CLIP \cite{pmlr-v139-radford21a} uses internet-scale language supervision to learn broader semantic concepts in the vision domain via contrastive pre-training over image-caption pairs.
In the Text-to-SQL task, where we perform similarity-based retrieval for ICL demonstrations, we face the same problem of representing asymmetric entities, namely the natural language question and the SQL query.
Some previous works \cite{ni-etal-2022-large} look to supervised fine-tuning where sufficient in-domain data is available, and others propose new objective functions that are optimized for the task or domain \cite{li2023angleoptimized}.
In this work, we devise a task-specific metric and perform weak supervised learning to align natural language with SQL using standard objective functions.
Our method effectively learns metadata-agnostic embeddings without masking in-domain data, customizing the base model vocabulary, rewriting questions, or using custom objective functions.

\section{Learning Metadata-Agnostic Text-to-SQL Embeddings}
\label{sec:sid}
We seek to learn aligned embeddings of natural language questions and SQL queries that are semantically meaningful for retrieving ICL demonstrations in the Text-to-SQL task.
We define a novel similarity metric in \S~\ref{subsec:edit_distance} based on a query edit distance heuristic.


\subsection{Query Edit Distance (QED)} \label{subsec:edit_distance}
Consider a dataset $D$ comprised of $n$ (database $d$, question $q$, query $s$) triplets $\left\{(q_i \mapsto s_i | d_i)\right\}^n$.
To measure the alignment between two exemplar pairs (i.e. $(q_i, s_i) \leftrightarrow (q_j, s_j)$), we take a query editing perspective and use a keyword matching heuristic that matches the structural similarity between the queries $s_i$ and $s_j$ as a proxy.
Since the same expression can be expressed in many different ways using different SQL keywords, we group keywords based on their semantic nature and assign weights to them according to their impact on query structure (see \autoref{tab:conditionals}).
For each group $k$, we obtain SQL keywords of the minimal insertions ($I_k$) and removals ($R_k$) to change $s_i$ to $s_j$.
\begin{equation}\label{eq:qed}
\textsc{QED} = \sum_k \Bigl(w_k\cdot \lvert I_k - R_k \rvert + w_r\cdot \min(I_k, R_k)\Bigl)
\end{equation}
Intuitively, any keyword in group $k$ that can be replaced with another keyword from the same group to make $s_i$ closer to $s_j$ comprises a small difference between the queries.
For example, replacing \texttt{MIN} with \texttt{MAX}.
Each replacement operation contributes a low score of $w_r = 0.2$ to $\operatorname{QED}$.
Additional keywords that need to be added, for example if an \texttt{ORDER BY} clause is missing, contribute a score of $w_k$ based on the significance of the group to the structure of the query.
Operations pertaining to table and column names are ignored (e.g. aliases) and will result in the same score.
This ensures the representations are agnostic to the underlying metadata of $d_i$ and focus more on the query structure.

\subsection{Dataset}
We construct an augmented training dataset for our encoder using the training set of \textbf{Spider}\footnote{We exclude the Yelp and additional training sets} (a large-scale cross-domain Text-to-SQL benchmark that covers over 200 databases) comprised of 7000 examples \cite{Yu&al.18c}.
Following \S~\ref{subsec:edit_distance} each question is paired with all queries in the dataset to generate a training dataset $\bar{D} = \{(q_i, s_j, l_{ij})\}^{49M}$, where $l_{ij}$ represents a measure of alignment between $q_i$ and $s_j$ based on $\operatorname{QED}(s_i, s_j)$.
However, we find $\bar{D}$ is unbalanced, with only  $\sim$1\% of the queries considered similar (i.e. QED score lower than 1).
We undersample $\bar{D}$ by comparing the euclidean distance ($\tau$) of generic embeddings\textsuperscript{\ref{footnote:titan}} of $q_i$ and $q_j$ and $l_{ij}$. 
To enhance the learning process, we sample more from the discordant group of exemplars, which we define as those with a large $\tau = \lVert q_i-q_j \rVert^2 $ and low $l_{ij}$, or vice versa.
The total size of the filtered training set is brought down $\sim$2.4M with majority of the exemplars representing some degree of discordance, see \autoref{tab:undersampling} for a complete breakdown.
Since we are interested in learning fine-grained representations, we cap $QED \leq 5$, and min-max normalize the scores $l_{ij} \in [0,1]$ such that a score of 0 represents the most distant exemplars and is compatible with the chosen loss function.
See \autoref{table:distance_score_example} for a list of labeled examples.
For a complete set of hyperparameters used to train the encoder, refer to \autoref{tab:emb_hyperparameters}.

\subsection{Loss Function}
\begin{wrapfigure}{R}{0.35\textwidth}
    \vspace{-1.2cm}
    \centering
    \includegraphics[width=\linewidth]{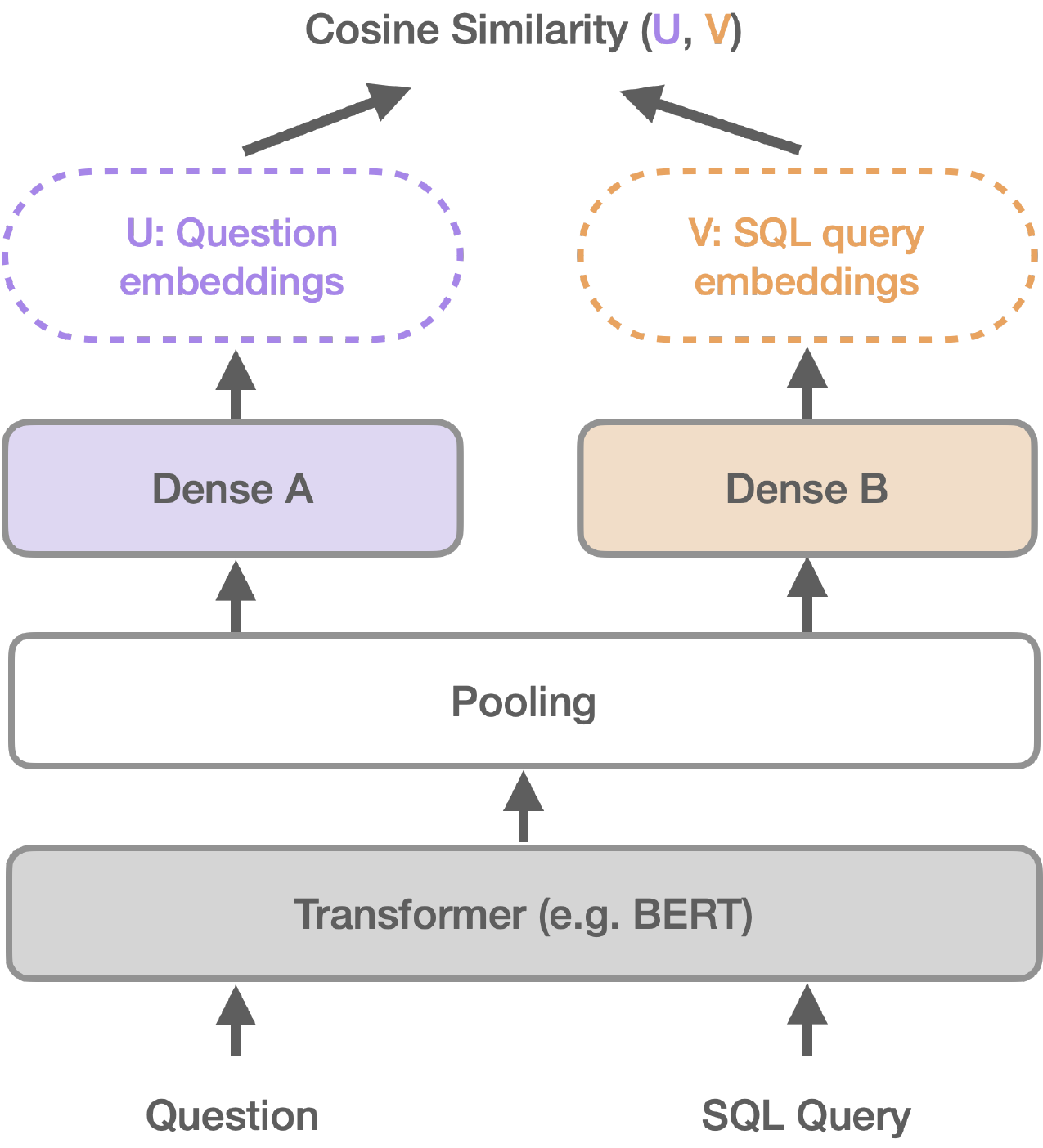}
    \caption{\textbf{Semi-asymmetric bi-encoder architecture.}. Parameters in the base transformer and the pooling layer are shared while two separate dense layers are trained to align question and SQL query embeddings, respectively.}
    \label{fig:architecture}
    \vspace{-0.1cm}
\end{wrapfigure}
The objective function used to train embedding models is usually selected based on the scale and nature of available training data.
For instance a contrastive loss \cite{ni-etal-2022-sentence, neelakantan2022text} such as Multiple Negatives Ranking Loss (MNRL) is used when pairwise data (e.g. images and text captions) is available, while a Triplet Loss is used when a neutral anchor can be compared to both positive and negative examples explicitly.
Normally Text-to-SQL datasets are comprised of (question, query) pairs, however a contrastive loss will likely not work well here as different questions might correspond to structurally similar queries in the same batch.
Hence, we create an augmented dataset with pseudo-labels based on \S~\ref{subsec:edit_distance} and train our encoder by minimizing the Cosine Similarity Loss ($\mathfrak{L}$),
where $\vec{q}_t$ and $\vec{s}_t$ are the question and query embeddings of the $t^{\text{th}}$ training example, respectively, $l_t$ is the alignment score derived using \autoref{eq:qed}, and $\mathrm{sim}(\vec{q}_t, \vec{s}_t)$ is the cosine similarity of the embeddings, defined in \autoref{eq:cosine}.
We choose the Cosine Similarity Loss rather than a Triple Loss as the latter does not distinguish distances between similar and dissimilar exemplars whereas the derived label $l_t$ does.
\begin{minipage}[t]{0.5\textwidth}
\begin{equation}
    \mathfrak{L}(\vec{q}_t, \vec{s}_t, l_t) = \parallel l_t - \mathrm{sim}(\vec{q}_t, \vec{s}_t) \parallel_2
\end{equation}
\end{minipage}%
\begin{minipage}[t]{0.5\textwidth}
\begin{equation}
\label{eq:cosine}
    \mathrm{sim}(\vec{q}_t, \vec{s}_t) = \frac{\vec{q}_t \cdot \vec{s}_t}{\parallel \vec{q}_t \parallel \parallel \vec{s}_t \parallel}
\end{equation}
\end{minipage}

\subsection{Architecture}
\label{subsec:arch}
Our embedding model is based on a semi-asymmetric bi-encoder architecture\footnote{Implemented using \url{https://www.sbert.net/}} (\autoref{fig:architecture}).
A pre-trained transformer\footnote{We use \texttt{bert-based-uncased}} serves as a common backbone to produce intermediate representations of either a question or a query using the same vocabulary.
A pooling layer connects the backbone to two separate dense layer heads, one for each embedding type.
Following common practice, we use mean pooling and \texttt{tanh} activation in the dense layers.
This architecture diverges from symmetric \cite{reimers-gurevych-2019-sentence} or asymmetric \cite{gillick2018end, lee2020learning}  sentence embedding models, where either all or no parameters are shared in the Siamese configuration, respectively (\autoref{fig:arch_choices}).
Given the declarative nature of SQL queries, we hypothesize that parameter sharing in the backbone benefits from the transfer learning abilities of pre-trained transformers and the separate embedding heads acts as a form of regularization during alignment.
In \S~\ref{subsec:ablation}, we validate this hypothesis.

\section{Experimental Setup}
\label{subsec:task}
Given an input question $q^{\prime}$, an LLM with frozen parameters $\theta$ and sampling parameters $\phi$ and a pool of $n$ demonstrations $ D = \left\{(q_i \mapsto s_i | d_i)\right\}^n$, the LLM will generate a SQL query $s^{\prime}$ by sampling from the distribution:
$
    s^{\prime} \sim \operatorname{LLM}_{\theta, \phi}\left(S\left(q^{\prime}, D\right) \oplus q^{\prime}\right)
$
where $S$ selects $k$ demonstrations from the pool $D$ based on $q^{\prime}$, and $\oplus$ formats the prompt with the inference exemplar alongside the demonstrations.
We provide sampling hyperparameters in \autoref{tab:llm_hyperparameters}.

\subsection{Evaluation Dataset and LLM}
Given the complexity of this task, we use both open and closed large, instruction-tuned models that exhibit competitive natural language understanding, reasoning, and coding abilities (Claude\footnote{\label{footnote:claude}Created by \href{https://www.anthropic.com/news/introducing-claude}{Anthropic}}, Mistral\footnote{\label{footnote:mistral}Created by \href{https://mistral.ai/news/mistral-large/}{Mistral AI}}, and Llama 3\footnote{\label{footnote:llama3}Created by \href{https://ai.meta.com/blog/meta-llama-3/}{Meta}}) to assess the generalizability of our approach, and evaluate their Text-to-SQL ICL capabilities on the \textbf{Spider} development (1034 instances) and test (2147 instances) sets.
Note, these two datasets and the training split used to train the embedding model do not have any database in common, which allows us to validate the generalizability and domain-agnostic property of our learned representations.
Using prompts like \autoref{lst:prompt} formatted suitably for each model, we insert the question and table metadata from each instance at inference time. 

\subsection{Demonstration Selection Methods}
We run a comprehensive list of experiments inspired by related work (many of which we are unable to reproduce exactly).
We include a zero-shot setting (no in-context demonstrations) as a baseline.
In total, we implement 5 different example selection methods to pick $k=8$ demonstrations (the number of examples is chosen based on an ablation, see \autoref{fig:num_demos}), namely:
\begin{itemize}[leftmargin=*]
\item\textbf{Random Sampling} Another common baseline which uniformly samples examples. 
\item\textbf{Question Similarity (QS)} The top-$k$ examples are selected based on the euclidean distance between embeddings\footnote{\label{footnote:titan} We use \texttt{amazon.titan-embed-g1-text-02}} of the test question and those of questions from the example pool. \citealt{liu-etal-2022-makes} adopted this approach using SBERT.
\item\textbf{Masked Question Similarity (MQS)} Same as QS, except information from the database metadata is masked in each question using a \texttt{[MASK]} token. Compared to our implementation, \citealt{guo2023case} use GPT-3.5 (\texttt{text-davinci-003}).
\item\textbf{Skills Similarity (SS)} An LLM first produces a summary description of the \emph{skills} required to solve any given exemplar, and then examples are retrieved based on the similarity of encoded\textsuperscript{\ref{footnote:titan}} skill descriptions.
We use the same prompt and 16 hand-crafted examples as \citealt{an-etal-2023-skill}.
\item\textbf{MARLO (\textit{this work})} Same as QS, except we use embeddings from our encoder described in \S~\ref{subsec:arch}. In \S~\ref{subsec:ablation} we also conduct an ablation using demonstrations retrieved with query embeddings instead of question embeddings.
\end{itemize}

\subsection{Metrics}
We evaluate all experiments using execution accuracy (EX), which measures the match between the execution outputs of the predicted and ground-truth SQL queries.
While the official Spider test suite (Seq-Eval) is used ubiquitously, it does not allow for the presence of redundant columns in the predicted query, which may be included depending on the LLMs query writing style.
To relax this requirement, we propose a modified execution accuracy (EX\textsuperscript{\textdagger}) which is based on Seq-Eval but considers all columnar permutations of the predicted query execution with the same number of columns as the ground-truth execution (\autoref{alg:ex}).

\section{Results \& Discussions}\label{subsec:results}

We present the main results from our experiments in \autoref{table:results}.
Overall, example selection with MARLO outperforms generic methods by +2.9 percentage points on average.
It also outperforms the next best method (SS) with significantly less inference latency as no additional LLM call is needed.

\begin{wraptable}{R}{0.55\textwidth}
\vspace{-1.2cm}
\caption{\textbf{Execution accuracy (\%) of example selection methods}. All experiments select $k=8$ examples. Results of the top performing method are in \textbf{bold}. MARLO mostly outperforms other methods, and is competitive otherwise.}
\begin{center}
\begin{small}
\begin{adjustbox}{max width=\textwidth}
\begin{tabular}{clccccc}
\toprule
\multirow{2}{*}{\textbf{LLM}} & \textbf{Selection} & \multicolumn{2}{c}{\textbf{Spider-dev}} && \multicolumn{2}{c}{\textbf{Spider-test}}\\ \cline{3-4} \cline{6-7}
 & \textbf{Method} & \textbf{EX} & \textbf{EX\textsuperscript{\textdagger}} && \textbf{EX} & \textbf{EX\textsuperscript{\textdagger}} \\
\midrule
\multirow{7}{*}{\shortstack[c]{Claude\\2.1}} & 0-shot & 67.5 & 78.5 && 69.0 & 80.0 \\
& Random & 72.2 & 79.4 && 75.1 & 82.2 \\
& QS & 73.5 & 80.5 && 75.1 & 80.5 \\
& MQS & 75.3 & 80.3 && 78.0 & 82.8 \\
& SS & 77.9 & 81.2 && 80.7 & 83.1 \\
& MARLO & \textbf{80.8} & \textbf{83.6} && \textbf{81.2} & \textbf{84.0} \\
\midrule
\multirow{6}{*}{\shortstack[c]{Mistral\\Large}} & 0-shot & 74.4 & 79.5 && 76.8 & 81.8 \\
& Random & 76.3 & 79.9 && 78.2 & 82.3 \\
& QS  & 77.7 & 79.8 && 79.4 & 82.9 \\
& MQS & 77.3 & 79.4 && 80.2 & 83.1 \\
& SS & 78.3 & 79.7 && 81.6 & \textbf{83.4} \\
& MARLO & \textbf{80.0} & \textbf{81.7} && \textbf{81.7} & 83.2 \\
\midrule
\multirow{6}{*}{\shortstack[c]{Llama 3\\70B\\Instruct}} & 0-shot & 77.0 & 81.5 && 75.6 & 81.3 \\
& Random & 79.0 & 81.8 && 78.6 & 81.8 \\
& QS & 81.5 & \textbf{83.1} && 78.6 & 80.7 \\
& MQS & 81.8 & 82.9 && 80.5 & 82.3 \\
& SS & 82.1 & 82.9 && \textbf{83.4} & \textbf{84.4} \\
& MARLO & \textbf{82.2} & 83.0 && 83.2 & 83.8 \\
\bottomrule
\end{tabular}
\end{adjustbox}
\end{small}
\end{center}
\label{table:results}
\vspace{-0.5cm}
\end{wraptable}
 
\subsection{Example Selection with MARLO} \label{subsec: MARLO_selection}
It is not surprising that MQS and SS outperform QS, as the former methods also exclude domain-specific information contained in the database metadata using either question masking or rephrasing, respectively.
MARLO, on the other hand, achieves the same objective without risking information loss or increasing inference latency and complexity (discussed further in \S~\ref{subsec:cost}).
It does so by leveraging fine-grained embeddings of natural language questions that are aligned to their corresponding SQL structure but still retain all relevant linguistic information.

\autoref{tab:retrieved-examples} further shows examples of different demonstrations retrieved using MARLO versus the other methods.
MARLO stands out by selecting demonstrations with structurally-alike queries and question intents across domains and a variety of question phrasings.
Its success in this task reveals that fine-grained task-specific embeddings can be used to select relevant but linguistically diverse demonstrations and, therefore, have the potential to help LLMs better comprehend natural language inputs of complex or multi-modal tasks during inference via ICL.
This ability is exemplified on difficult exemplars, where MARLO with both Claude 2.1 and Mistral Large significantly outperform other methods by +4-7\%pt. (see \autoref{fig:breakdown}).
 
\subsection{LLM Text-to-SQL Capability} \label{subsec:claude_cap}
\paragraph{EX vs EX\textsuperscript{\textdagger}}
As shown in \autoref{table:results}, the accuracy scores are systematically higher yet more stable when measured with EX\textsuperscript{\textdagger} regardless the LLM chosen. 
This effect is more pronounced in Claude 2.1 than other LLMs.
Upon investigation, we find that SQL queries produced by Claude 2.1 tend to order numerical columns before categorical columns and include additional redundant columns.
This explains why its EX is significantly lower, while EX\textsuperscript{\textdagger} is on par with other LLMs.
We hypothesize an LLM's query writing style, and therefore sensitivity under EX, is attributed to differences in training data and instruction tuning.

\paragraph{ICL \& Llama 3 Instruct}
We observe that Llama 3 Instruct behaves differently to the other LLMs.
It exhibits competitive zero-shot performance but does not benefit as much from ICL, regardless of the selection method used.
It is possible that its instruction-tuning procedure trades off Text-to-SQL ICL ability as an ``alignment-tax'' \cite{ouyang2022training}, or is unable to benefit from proprietary data.

\paragraph{Comparison with GPT 4}
Despite the inconsistent use of sampling hyperparameters, prompts, and number of examples reported in literature, we list the execution accuracy reported by other studies using GPT 4 for comparison in \autoref{table:gpt4}.
EX\textsuperscript{\textdagger} is computed for comparison when possible (i.e. works are reproducible or their predictions are shared publicly).
We observe a similar trend of performance improvement across the selection methods with GPT 4 as with Claude 2.1 and Mistral Large.
Notably, MARLO using Claude 2.1 is competitive with DAIL-SQL\footnote{\label{footnote:sota}We consider DAIL-SQL state-of-the-art (SOTA)} using GPT 4 on Spider-dev (EX\textsuperscript{\textdagger}: 83.6 vs. 83.3, respectively).

\subsection{Ablation Study} \label{subsec:ablation}
\begin{wrapfigure}{R}{0.5\textwidth}
    \vspace{-0.6cm}
    \centering
    \includegraphics[width=\linewidth]{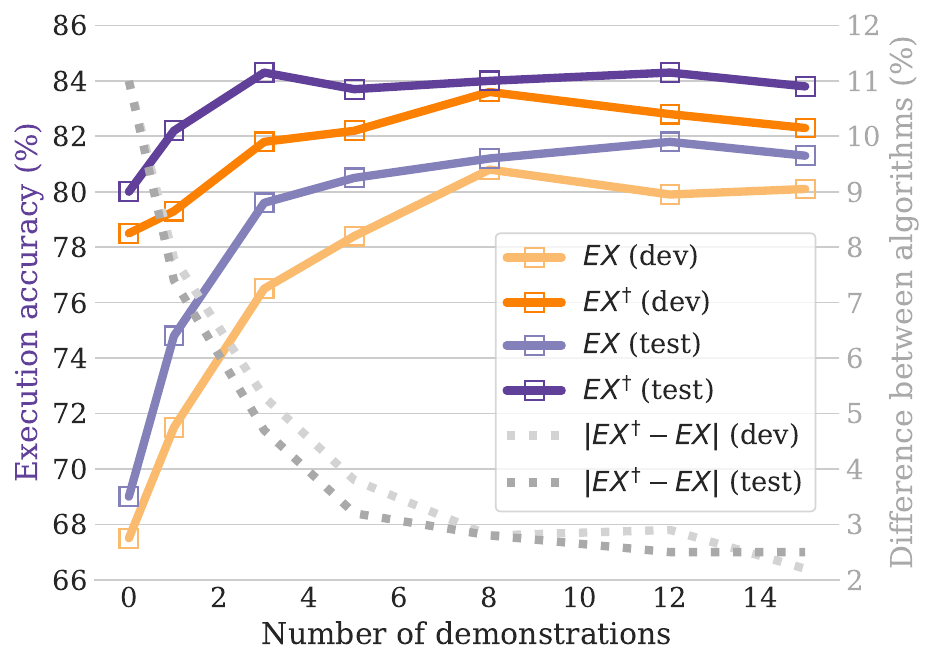}
    \caption{\textbf{Execution accuracy (\%) of MARLO for various numbers of selected demonstrations.} Performance initially increases and then plateaus as more demonstrations are included in the context, implying in-context learning scaling limitations for this task.
    }
    \label{fig:num_demos}
    \vspace{-1em}
\end{wrapfigure}

We perform four ablations to study our proposed retrieval process, illustrated in \autoref{tab:ablations} and \autoref{fig:num_demos}.
\begin{wraptable}{R}{0.55\textwidth}
\vspace{-1.3em}
\caption{\textbf{Execution accuracy (\%) of MARLO ablations}. Ablation (\subref{tab:ablation_a}) explores the effect of parameter sharing in the encoder architecture, (\subref{tab:ablation_b}) compares the chosen objective function to Multiple Negatives Ranking Loss (MNRL), and (\subref{tab:ablation_c}) compares end-task performance for various retrieval recipes. In (\subref{tab:ablation_c}) a relative number of unique demonstration selected during each evaluation ($\mathfrak{R}$) is also reported. All studies use Claude 2.1.}
\centering
\begin{subtable}[h]{\linewidth}
\begin{center}
\caption{retriever dual-encoder architectures}
\begin{small}
\begin{tabular}{cccccc}
\toprule
\multirow{2}{*}{\textbf{Architecture}}  & \multicolumn{2}{c}{\textbf{Spider-dev}} && \multicolumn{2}{c}{\textbf{Spider-test}}\\ \cline{2-3} \cline{5-6}
& \textbf{EX} & \textbf{EX\textsuperscript{\textdagger}} && \textbf{EX} & \textbf{EX\textsuperscript{\textdagger}} \\
\midrule
Symmetric & 78.6 & 81.4 && 80.9 & 83.6 \\
\textbf{Semi-asymmetric} & \textbf{80.8} & \textbf{83.6} && 81.2 & \textbf{84.0} \\
Asymmetric & 77.6 & 81.6 && \textbf{81.7} & \textbf{84.0} \\
\bottomrule
\end{tabular}
\end{small}
\label{tab:ablation_a}
\end{center}
\end{subtable}
\begin{subtable}[h]{\linewidth}
\begin{center}
\caption{retriever training objective functions}
\begin{small}
\begin{tabular}{cccccc}
\toprule
\multirow{2}{*}{\textbf{Loss function}}  & \multicolumn{2}{c}{\textbf{Spider-dev}} && \multicolumn{2}{c}{\textbf{Spider-test}}\\ \cline{2-3} \cline{5-6}
& \textbf{EX} & \textbf{EX\textsuperscript{\textdagger}} && \textbf{EX} & \textbf{EX\textsuperscript{\textdagger}} \\
\midrule
MNRL & 75.8 & 81.6 && 79.1 & 82.9 \\
\textbf{Cosine Similarity} & \textbf{80.8} & \textbf{83.6} && \textbf{81.2} & \textbf{84.0} \\
\bottomrule
\end{tabular}
\end{small}
\label{tab:ablation_b}
\end{center}
\end{subtable}
\begin{subtable}[h]{\linewidth}
\begin{center}
\caption{linearly combined question/query embeddings}
\begin{small}
\begin{tabular}{cccccccc}
\toprule
\textbf{Embed.} & \multicolumn{3}{c}{\textbf{Spider-dev}} && \multicolumn{3}{c}{\textbf{Spider-test}}\\ \cline{2-4} \cline{6-8}
\textbf{Weight} & \textbf{EX} & \textbf{EX\textsuperscript{\textdagger}} & \textbf{$\mathfrak{R}$} && \textbf{EX} & \textbf{EX\textsuperscript{\textdagger}} & \textbf{$\mathfrak{R}$} \\
\midrule
Query & 76.3 & 82.3 & .28 && 79.2 & \textbf{84.3} & .38 \\
50 / 50 & 79.0 & 82.2 & .45 && \textbf{81.3} & 84.1 & .62 \\
70 / 30 & 79.0 & 82.9 & .55 && 81.0 & 83.4 & .77 \\
Question & \textbf{80.8} & \textbf{83.6}  & .71 && 81.2 & 84.0 & 1 \\
\bottomrule
\end{tabular}
\end{small}
\label{tab:ablation_c}
\end{center}
\end{subtable}
\label{tab:ablations}
\vspace{-0.5cm}
\end{wraptable}
\paragraph{Encoder Architecture}
As discussed in \S~\ref{subsec:arch}, the architecture of our encoder deviates from that of common symmetric (e.g. SBERT) and asymmetric (e.g. CLIP) bi-encoders.
In \autoref{tab:ablation_b} we explore the effect of parameter sharing by training either a single encoder to embed both questions and queries, a separate encoder for each, or separate output layers that process representations from a shared backbone.
See \autoref{fig:arch_choices} for the three different architectures explored.
We find the semi-asymmetric architecture is more capable of learning expressive and aligned representations than the symmetric or asymmetric alternatives.
We expect parameter sharing in the backbone benefits from the transfer learning abilities of large pre-trained language models and helps build alignment between latent representations of question and queries from the same parameter space.
This idea has been explored and validated by \citealt{dong2022exploring} in the context of question-answer retrieval.
Moreover, separation of parameters in the later layers helps enforce alignment in the backbone independent of the way the questions or queries are encoded, which explains why the asymmetric architecture performs slightly better than a symmetric one.

\paragraph{Encoder Loss Function}
Retrievers are often trained with contrastive loss functions when pairwise data (such as questions and queries) is available.
Here, we compare the effectiveness of using more fine-grained pseudo-labels derived from the metric we propose in \S~\ref{subsec:edit_distance} with the Multiple Negatives Ranking Loss (MNRL, \citealt{henderson2017efficient}).
MNRL computes the cross entropy of all possible question-query combinations in a batch, where pairs of questions and their ground-truth queries are assigned a positive label and all other pairs a negative label.
Due to the impact of the batch size on the loss functions, we adjust the training hyperparameters for MNRL to ensure sufficient coverage of both positive and negative pairs (see \autoref{tab:emb_hyperparameters}).
As \autoref{tab:ablation_b} shows, cosine similarity using pseudo-labels derived using QED leads to better end-task performance than MNRL.
This is because QED scores provide better signal for supervision than the binary scores used in MNRL and are less noisy than the latter.
It is probable that false negatives exist in batch for questions that are dissimilar but have similar corresponding queries, and vice versa.
Nevertheless, the results using MNRL are either on par or better than those achieved with other example selection methods, highlighting the utility of aligned embeddings.

\paragraph{Embedding Alignment \& Retrieval Recipe}
\label{par:embegging-alignment}
Unlike previous work (e.g. \citealp{gao2023texttosql}), our question and query embeddings can be used interchangeably without the additional overhead of predicting a preliminary query because they are closely aligned.
\autoref{tab:ablation_c} shows that selecting demonstrations for a given test question based on its semantic similarity to candidate questions or queries results in comparable performance on EX\textsuperscript{\textdagger}.
Nevertheless, question embeddings are more expressive than their query counterparts as the former leads to a greater number of unique demonstrations selected during the evaluation ($\mathfrak{R}$).
Therefore, it appears to be more beneficial when including the exact number of output fields in the predicted query is required (i.e. EX).
Intuitively, a natural language question can be written in many more ways than its corresponding SQL query, which explains why their respective embeddings are aligned but not equally expressive.
Although linear combinations of question and query embeddings does not result in obvious gains, we believe further exploration of the compositionality and factorizaiton of these embeddings \cite{trager2024linear} can help boost performance and support their broader application.

\paragraph{Number of Selected Demonstrations}
Overall, we observe Text-to-SQL performance initially increases then plateaus as we increase the number of in-context demonstrations (\autoref{fig:num_demos}).
It is likely the case that selecting demonstrations based solely on semantic similarity has its limits, as the information gain of each new demonstration added to the context saturates.
Perhaps an ensemble of example selection methods that optimize for other aspects might be beneficial in large-context settings.
In addition, EX is systematically lower than EX\textsuperscript{\textdagger}, and this effect is more pronounced when fewer demonstrations are used in context.
Based on the asymptotic difference between evaluation algorithms in \autoref{fig:num_demos}, one of the first patterns LLMs learn from the in-context demonstrations is an understanding of what (and the exact number of) fields to include in the predicted query.
Since this pattern solely relies on question understanding as apposed to query writing abilities, our interpretation is that the LLM uses its inherent (zero-shot) SQL understanding to inform question understanding.
This corroborates recent work by \citealp{wang-etal-2023-label} who argue demonstration outputs serve as anchors through which information flows from the demonstration inputs in the shallow layers of the model \cite{min2022rethinking, dziri2023faith}.
We expect these anchors represent latent interpretations of the questions in the context of their corresponding queries, and in the deeper layers of the model, the relevant representations inform the query generation process for the test exemplar.

\section{Conclusions}
In this work we explore the problem of selecting demonstration examples to improve ICL capabilities of LLMs on the Text-to-SQL task.
Given the disjointed distributions of natural language questions and SQL queries, general-purpose internet-scale encoders are generally not well-suited to retrieve semantically similar demonstrations.
Therefore, we propose a novel approach, MARLO, that trains a bi-encoder to align the representations of natural language questions and SQL queries according to their underlying intent.
Via weak supervision from a metadata-agnostic similarity label, MARLO selects ICL demonstrations that enables LLMs to excel in generating queries.
Not only are our results competitive with the state-of-the-art, MARLO is also more efficient and effective than previous domain masking techniques.
Our ablations reveal that the selections it provides are semantically relevant for the task yet linguistically unconstrained.
MARLO's success suggests that fine-grained task-specific embeddings have the potential to enhance LLMs in complex or multi-modal ICL settings.

\bibliography{main}

\newpage
\appendix
\onecolumn
\section{Appendix}
\label{sec:appendix}

\subsection{Limitations}
\label{sec:limitations}
\subsubsection{Open vs. Closed Models}
Closed foundation models significantly outperform open models across a broad range of complex reasoning tasks and offer certain capabilities to which open models have not caught up.
However, two major limitations stand out when using closed models for research.
First, the cost and throughput can be prohibitively expensive and slow, respectively.
Second, the lack of interoperability of these models and transparency concerning their training data and methods creates a restrictive research environment in which reproducing and benchmarking against prior works becomes challenging.
Hence, we are not able to perform a direct and thorough comparison of Text-to-SQL capabilities between additional LLMs (e.g. GPT 4) by keeping hyperparameters, prompting technique and example selection method in a controlled manner.
Moreover, it is difficult to reason about performance differences between open and closed models as we are unable to compare parameter counts, training data, or fine-tuning and alignment methods.

\subsubsection{Automatic Query Generation}
While SQL generation solutions built using probabilistic models might appeal to database management and information extraction business use cases, the risk of hallucination and catastrophic customer impact remains to be explored.
By observing the upper-bound performance in \autoref{table:results}, we see there is still ample room for improvement on the robustness of LLMs across benchmarks and domains.

\subsection{MARLO Encoder Training}
\subsubsection{Hyperparamters}
In \autoref{tab:emb_hyperparameters} we provide the hyperparamters used for all MARLO encoder experiments. All experiments use a NVIDIA A10G Tensor Core GPU.
\begin{table}[ht]
\caption{\textbf{Hyperparameters of training our customized embedding model}.}
\begin{center}
\begin{tabular}{lll}
\toprule
\textbf{Hyperparameter} & \textbf{Cosine Similarity} & \textbf{MNRL}\\
\midrule
maximum input sequence length & 256 & 256\\
fixed output size & 128 & 128 \\
batch size & 32 & 32 \\
Optimizer & \texttt{AdamW} & \texttt{AdamW} \\
learning rate schedule & linear warm-up & linear warm-up \\
maximum learning rate & $2\times 10^{-5}$ & $2\times 10^{-5}$ \\
\# of learning rate warm up steps & 10000 & 100 \\
weight decay for model parameters & 0.01 & 0.01 \\
maximum gradient normalization & 1 & 1 \\
\# of epochs & 2 & 15 \\
\bottomrule
\end{tabular}
\end{center}
\label{tab:emb_hyperparameters}
\end{table}

\subsubsection{Sampling Discordant Questions \& Queries}
\autoref{tab:undersampling} displays details about how we undersample our original 49M pairs of questions and queries to the final training set used in MARLO. 
We sample the given number of exemplars (from the total per category) according to QED score and euclidean distance between embedded questions $q_i$ and $q_j$ criteria outlined in the table. 
To facilitate better learning of the encoder, we intentially select more examples from discordant categories.
\begin{table}[h]
\caption{\textbf{Under-sampling of augmented spider training set}. We reduce the training dataset from 49M to $\sim$2.4M.}
\begin{center}
\begin{tabular}{cccc}
\toprule
$\tau = \lVert q_i-q_j \rVert^2 $ & $1 < l_{ij}$ & $1 \leq l_{ij} < 3$ & $3 \leq l_{ij}$\\
\midrule
$\tau \geq 300$ & 600K ($\sim$5.4M) & 600K ($\sim$10M) & 100K ($\sim$32.7M)\\
$\tau < 300$ & 170K (170K) & 150K (150K) & 800K (800K)\\
\bottomrule
\end{tabular}
\end{center}
\label{tab:undersampling}
\end{table}

\subsubsection{Similarity via Query Edit Distance}
This section provides the details about how we compute Query Edit Distance (QED) score based on the differences in SQL keywords that appear in insert and remove operations when two queries are compared.
\begin{table}[ht]
\caption{\textbf{SQL keyword group weights.} For groups with multiple keywords, a universal weight $0.3$ is used. Since these groups can have non-zero counts of both insertions and removals, they are considered to lead smaller within-group distances and are assigned a weight of $0.2$. See \autoref{eq:qed} for details.
While this list is complete for our given dataset, it might not be complete for all possible queries and dialects. Additional keywords would have to be included appropriately.}
\begin{center}
\begin{tabular}{llc}
\toprule
\textbf{Group ($k$)} & \textbf{SQL Keywords} & \textbf{$w_k$}\\
\midrule
Aggregation & \texttt{COUNT}, \texttt{AVG}, \texttt{SUM}, \texttt{MIN}, \texttt{MAX} & \multirow{4}{*}{0.3}\\
Comparison & \texttt{EQ}, \texttt{NEQ}, \texttt{LIKE}, \texttt{GT}, \texttt{GTE}, \texttt{LT}, \texttt{LTE}, \texttt{BETWEEN}, \texttt{IN} & 
\\
Composition & \texttt{AND}, \texttt{OR} & \\
Arithmetic & \texttt{ADD}, \texttt{SUB} & \\
\midrule
\multirow{12}{*}{--} & \texttt{LIMIT} & 0.1\\
& \texttt{DISTINCT} & 0.2\\
& \texttt{WHERE} & 0.5\\
& \texttt{HAVING} & 0.7\\
& \texttt{GROUP} & 0.6\\
& \texttt{ORDER} & 0.6\\
& \texttt{JOIN} & 3.0\\
& \texttt{SELECT} & 3.0\\
& \texttt{SUBQUERY} & 4.0\\
& \texttt{EXCEPT} & 4.0\\
& \texttt{UNION} & 3.0\\
& \texttt{INTERSECT} & 3.5\\
\bottomrule
\end{tabular}
\end{center}
\label{tab:conditionals}
\end{table}
Since SQL keywords are grouped based on their influence on the overall query structure (\autoref{tab:conditionals}), we consider two different queries closer in edit distance when their differences are within the same compared. See \autoref{eq:qed} for the formula of QED computation.

\subsubsection{QED Example}
\autoref{table:distance_score_example} shows an example of how QED scores and corresponding similarity labels look like for a single question paired with four different possible queries. 
The top row is the ground truth query of the question so the QED score is 0 and similarity score is 1.
From top to bottom, we see an increase in QED, representing the queries become more and more irrelevant to the question asked.
\begin{table*}[ht]
\caption{\textbf{Examples of QED score and similarity score labeling}. The scores are impacted by the dissimilar SQL structure but not by the domain information.}
\small
\begin{center}
\begin{tabularx}{\textwidth}{ssbdd} 
\toprule
\textbf{Question} & \textbf{Ground-Truth Query} & \textbf{Possible Query} & \textbf{QED Score} & \textbf{$l_{ij}$}\\
\midrule
\multirow{11}{=}{How many heads of the departments are older than 56?} & \multirow{11}{=}{\texttt{SELECT count(*) FROM head WHERE age > 56}} & \texttt{SELECT count(*) FROM head WHERE age > 56} & 0 & 1 \\[0.5cm]
& & \texttt{SELECT count(*) FROM professor WHERE prof\_high\_degree = 'Ph.D.'} & 0.2 & 0.96 \\[0.5cm]
& & \texttt{SELECT major,  count(*) FROM Student GROUP BY major} & 1.4 & 0.72 \\[0.5cm]
& & \texttt{SELECT DISTINCT T1.age FROM management AS T2 JOIN head AS T1 ON T1.head\_id = T2.head\_id WHERE T2.temporary\_acting = 'Yes'} & 5 & 0 \\
\bottomrule
\end{tabularx}
\end{center}
\label{table:distance_score_example}
\end{table*}

\subsubsection{Bi-encoder Architectures}
\begin{figure}
    \centering
    \begin{subfigure}{.3\textwidth}
        \centering
        \includegraphics[width=0.9\linewidth]{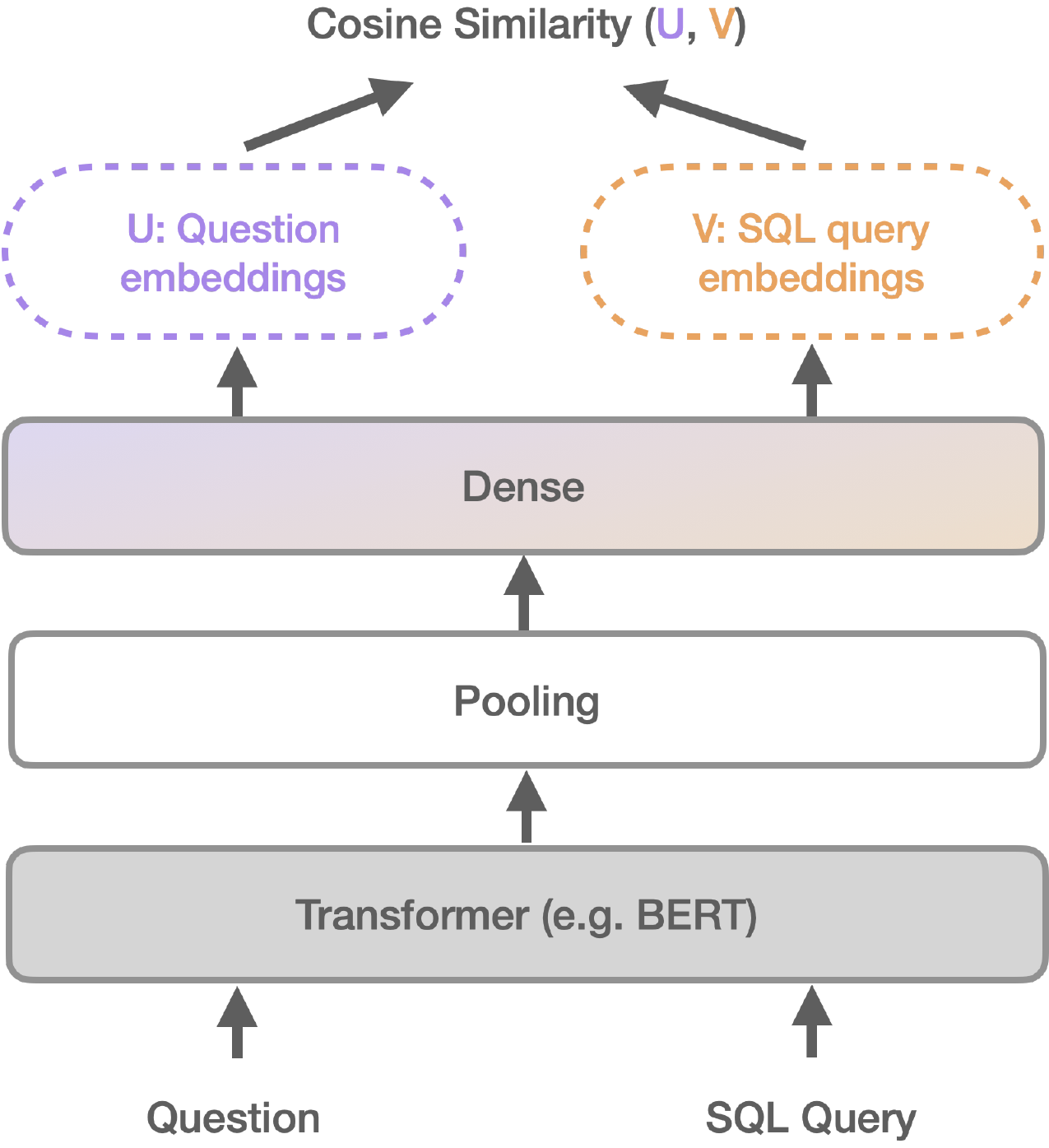}  
        \caption{Symmetric}
        \label{sym_arch}
    \end{subfigure}
    \begin{subfigure}{.3\textwidth}
        \centering
        \includegraphics[width=.9\linewidth]{figs/architecture-n.pdf}  
        \caption{Semi-asymmetric \textit{(this work)}}
        \label{arch}
    \end{subfigure}
    \begin{subfigure}{.3\textwidth}
        \centering
        \includegraphics[width=.9\linewidth]{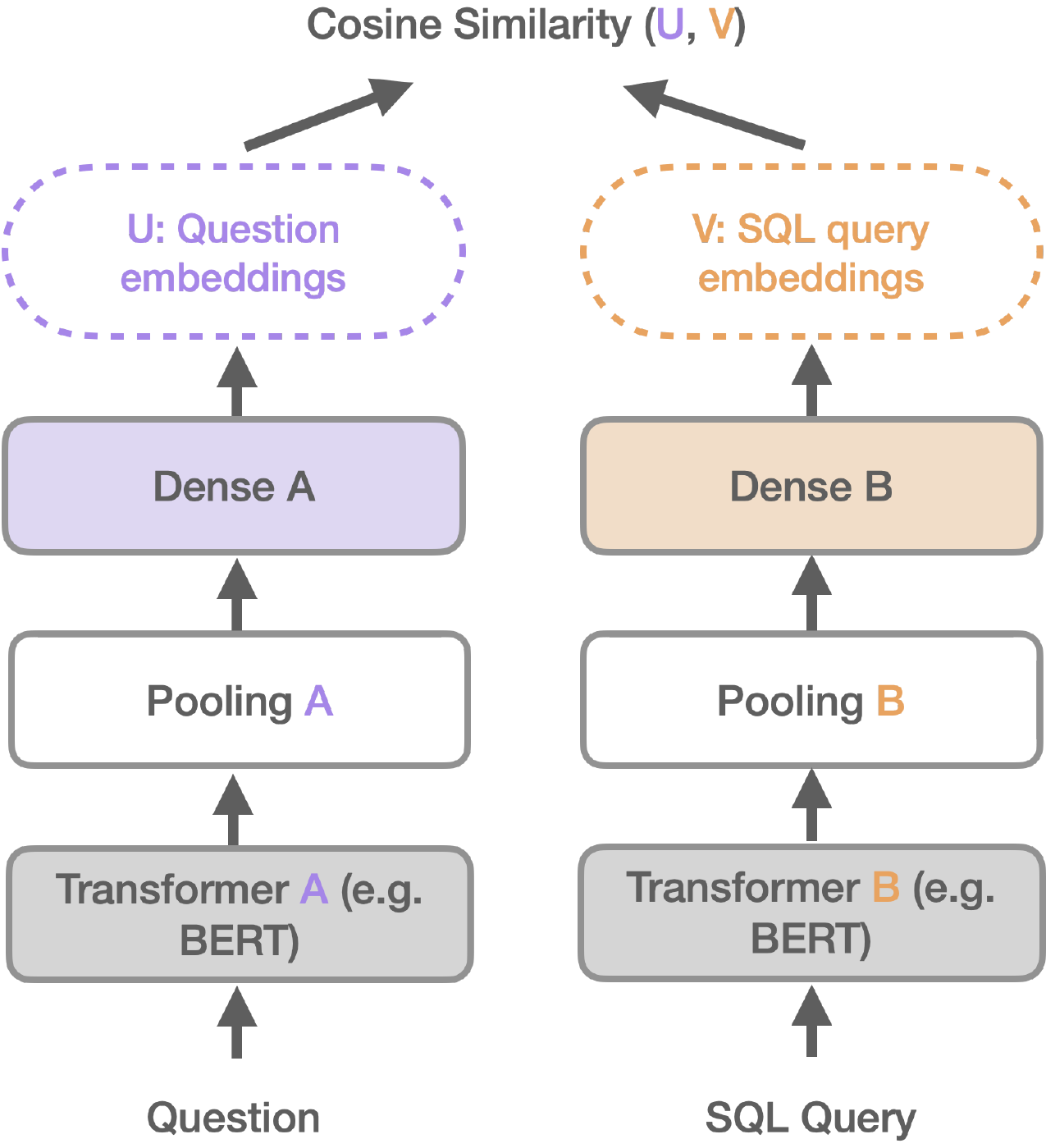}  
        \caption{Asymmetric}
        \label{asym_arch}
    \end{subfigure}
    \caption{\textbf{Architectures choices for bi-encoders.} A symmetric architecture (a) have parameters shared in all three modules while an asymmetric architectures (c) does not share any layer between the two towers. Our work adopted a semi-asymmetric structure where a common backbone transformer and pooling layer are shared, but dense layers are separated.}
    \label{fig:arch_choices}
\end{figure}

\autoref{fig:arch_choices} shows three different architecture choices for bi-encoders. Symmetric architectures are commonly used in literature (e.g. SBERT) while asymmetric architectures are also found effective in use cases where the input and output have different lengths or distributions (e.g. document retrieval based on questions). 
Our encoder is trained from a semi-asymmetric architecture (\autoref{arch}) where the backbone transformer and pooling layers are shared but dense layers are independent.

\subsection{Query Generation \& Evaluation}
\subsubsection{Decoding Hyperparamters}
\autoref{tab:llm_hyperparameters} provides the configuration used to predict SQL queries. For experiments that require multiple predicted queries (i.e. self-consistency and upper bound) we use a non-zero temperature of 0.7.
\begin{table}[ht]
\caption{\textbf{Hyperparameters for SQL query generation}.}
\begin{center}
\begin{tabular}{ll}
\toprule
Hyperparameter & Value \\
\midrule
\multirow{2}{*}{temperature} & 0.7 (when sampling multiple queries)\\
& 0  (otherwise)\\
top $k$ & 400 \\
top $p$ & 1 \\
maximum \# of sampling tokens & 1000 \\
stop sequence & \texttt{``</sql>''} \\
\bottomrule
\end{tabular}
\end{center}
\label{tab:llm_hyperparameters}
\end{table}

\subsubsection{Prompts for Claude in SQL Query Generation}
Claude is trained to generate text as an AI assistant in a dialogue with a human user, so the prompt includes \texttt{Human:} and \texttt{Assistant:} prefixes to indicate the context.
Claude also prefers to work with XML tags which segment different parts of the prompt and help us parse the outputs reliably.
\autoref{lst:prompt} is the prompt format we use for Claude to perform the Text-to-SQL task throughout all experiments.
\begin{listing}[ht]
\caption{\textbf{Prompt for Claude}. For zero-shot experiments, the example block is removed. Entities enclosed in \{\} are input elements inserted at inference time. \texttt{\{demonstration\}} are constructed with the corresponding metadata, question, and query of the selected examples following the same formatting below. \texttt{\{table\_schemas\}} are the \texttt{CREATE} statements for each table in the database and \texttt{\{row\_inspections\}} provide a single sample row for each table in YAML format.}
\begin{small}
\begin{minted}[mathescape,
linenos=False,
bgcolor=LightGray,
gobble=2,
breaklines]{yaml}
  Human: Paying careful attention to the table and column names in the given metadata, provide a correct {dialect} query to answer the given question. Enclose your query in '<sql></sql>' XML tags.
  # example block
  Here are some examples:
  <example>{demonstration}</example>
  .
  .
  .
  # metadata block
  Metadata:
  <metadata>
  {table_schemas}
  {row_inspections}
  </metadata>
  # input question
  Question: {question}
  Assistant: SQL Query: <sql>
\end{minted}
\end{small}
\label{lst:prompt}
\end{listing}
In \autoref{lst:example-prompt} we provide a complete, formatted prompt with a single in-context demonstration and real sample data.

\subsubsection{Modified Execution Accuracy Algorithm}
In \autoref{alg:ex}, we presents our modified execution accuracy evaluation method which allows for redundant columns to be included in predicted queries.
\begin{algorithm}[ht]
   \caption{\textbf{Modified Execution Accuracy (EX\textsuperscript{\textdagger})} compares ground-truth query results to columnar permutations (of the same length) of the predicted query results. $P(a,b)$ denotes all permutations of list $a$ with length $b$, and $\left[\cdot\right]$ columnar indexing.}
   \label{alg:ex}
\begin{algorithmic}
\STATE {\bfseries Input:} execution engine $E$, database $D$, gold query $g$, predicted query $p$
\STATE Compute gold exec. results $r_g := E(D, g)$
\STATE Compute pred. exec. results $r_p := E(D, p)$
\STATE $accurate = False$
\FOR{ $c_{p_i}$ \bfseries in $P\left(cols\left(r_p\right),\left|cols\left(r_g\right)\right|\right)$}
\IF{$r_g$ {equals} $r_p\left[c_{p_i}\right]$}
\STATE $accurate = True$
\STATE break
\ENDIF
\ENDFOR
\end{algorithmic}
\end{algorithm}

\newpage
\subsection{Additional Data for Result Analysis}

\subsubsection{Median QED Achieved by Different Selection Methods}
\autoref{table:mqed} shows the median QED scores between gold queries and those of the selected demonstrations from different selection methods we experiment.
Our encoder minimizes QED and helps MARLO select examples that leads to better end task performance.

From the median QED scores between gold queries and those of the selected demonstrations (\autoref{table:mqed}), we observe how QS, MQS and SS are able to retrieve examples that have lower QED scores than random selection, which is correlated with their better end task performance.
And that our encoder indeed minimizes QED while selecting examples for MARLO which further enhances the performance.
\begin{table}
\caption{\textbf{Median Query Edit Distance ($M(\operatorname{QED})$) between selected and ground truth examples for selection methods on Spider}. $M(\operatorname{QED})$ generalizes to new domains as it is minimized considerably using MARLO compared to other domain masking methods.
} 
\begin{center}
\begin{small}
\begin{adjustbox}{max width=\textwidth}
\begin{tabular}{lcc}
\toprule
\multirow{2}{*}{\textbf{Selection Method}} & \multicolumn{2}{c}{$M(\operatorname{QED})$} \\ \cline{2-3}
& \textbf{Spider-dev} & \textbf{Spider-test} \\
\midrule
Zero-shot Baseline & -- & -- \\
Random & 53.4 & 62.8 \\
Question Similarity (QS) & 50.7 & 51.0 \\
Masked Question Similarity (MQS) & 46.0 & 50.3 \\
Skills Similarity (SS) & 38.2 & 38.1 \\
MARLO (\emph{ours}) & 0.2 & 0.2 \\
\bottomrule
\end{tabular}
\end{adjustbox}
\end{small}
\end{center}
\label{table:mqed}
\end{table}

\subsubsection{Comparison of Retrieved Examples}
In \autoref{tab:retrieved-examples} we compare retrieved natural language question and SQL query pairs for a single inference exemplar across  selection methods. 
The metadata-agnostic property of MARLO is clearly demonstrated: structurally similar SQL queries and semantically similar questions are selected, under question phrasing and domains shifts.
We expect this helps the LLM understand the given question conditioned on the anticipated query structure.
In contrast across the other methods, the question phrasing may be diverse but the query structural less similar than the target, or the question and queries are too diverse but perhaps semantically ``close'' to the word \textbf{``conference''}.
\begin{scriptsize}
\begin{longtable}{ p{0.07\textwidth} p{0.34\textwidth} p{0.42\textwidth} p{0.03\textwidth} }
\toprule
\textbf{Selection Method} & \textbf{Retrieved Question} & \textbf{Retrieved Query} & \textbf{QED} \\
\midrule
\multirow{9}{=}{MARLO} & Return the different countries for artists. & \texttt{SELECT DISTINCT country FROM artist} & 0.0 \\
& Show all distinct building descriptions. & \texttt{SELECT DISTINCT building\_description FROM Apartment\_Buildings} & 0.0 \\
& What are the different film Directors? & \texttt{SELECT DISTINCT Director FROM film} & 0.0 \\
& Show all distinct lot details. & \texttt{SELECT DISTINCT lot\_details FROM LOTS} & 0.0 \\
& Give the distinct headquarters of manufacturers. & \texttt{SELECT DISTINCT headquarter FROM manufacturers} & 0.0 \\
& What are all the different book publishers? & \texttt{SELECT DISTINCT publisher FROM book\_club} & 0.0 \\
& List all different genre types. & \texttt{SELECT DISTINCT name FROM genres;} & 0.0 \\
& List the distinct director of all films. & \texttt{SELECT DISTINCT Director FROM film} & 0.0 \\
\midrule \multirow{13}{=}{Skill Similarity} & Show all transaction types. & \texttt{SELECT DISTINCT transaction\_type FROM Financial\_Transactions}  & 0.0 \\
& Show all video game types. & \texttt{SELECT DISTINCT gtype FROM Video\_games} & 0.0 \\
& What are the different card type codes? & \texttt{SELECT DISTINCT card\_type\_code FROM Customers\_Cards} & 0.0 \\
& What are the different types of player positions? & \texttt{SELECT count(DISTINCT pPos) FROM tryout} & 0.3 \\
& What are the different cities where people live? & \texttt{SELECT DISTINCT T1.city FROM addresses AS T1 JOIN people\_addresses AS T2 ON T1.address\_id = T2.address\_id} & 100.0 \\
& Show all product sizes. & \texttt{SELECT DISTINCT product\_size FROM Products} & 0.0 \\
& What document status codes do we have? & \texttt{SELECT document\_status\_code FROM Ref\_Document\_Status;} & 0.2 \\
& Find all the vocal types. & \texttt{SELECT DISTINCT TYPE FROM vocals} & 0.0 \\
\midrule
\multirow{12}{=}{Masked Question Similarity} & What are the different allergy types? & \texttt{SELECT DISTINCT allergytype FROM Allergy\_type} & 0.0 \\
& What are the different card type codes? & \texttt{SELECT DISTINCT card\_type\_code FROM Customers\_Cards} & 0.0 \\
& What are the different product sizes? & \texttt{SELECT DISTINCT product\_size FROM Products} & 0.0 \\
& What are the different product colors? & \texttt{SELECT DISTINCT product\_color FROM Products} & 0.0 \\
& What are the numbers of constructors for different nationalities? & \texttt{SELECT count(*) ,  nationality FROM constructors GROUP BY nationality} & 1.1 \\
& What are the number of different course codes? & \texttt{SELECT count(DISTINCT crs\_code) FROM CLASS} & 0.3 \\
& What are the different film Directors? & \texttt{SELECT DISTINCT Director FROM film} & 0.0 \\
& What are the different membership levels? & \texttt{SELECT count(DISTINCT LEVEL) FROM member} & 0.3 \\
& What are the numbers of wines for different grapes? & \texttt{SELECT count(*) ,  Grape FROM WINE GROUP BY Grape}  & 1.1 \\
\midrule
\multirow{23}{=}{Question Similarity} & What is the primary conference of the school that has the lowest acc percent score in the competition? & \texttt{SELECT t1.Primary\_conference FROM university AS t1 JOIN basketball\_match AS t2 ON t1.school\_id  =  t2.school\_id ORDER BY t2.acc\_percent LIMIT 1}  & 100.0\\
& What are the enrollment and primary conference for the university which was founded the earliest? & \texttt{SELECT enrollment ,  primary\_conference FROM university ORDER BY founded LIMIT 1}  & 0.9 \\
& What are the names of all teams? & \texttt{SELECT Name FROM Team} & 0.2 \\
& What are the names of colleges that have two or more players, listed in descending alphabetical order? & \texttt{SELECT College FROM match\_season GROUP BY College HAVING count(*)  >=  2 ORDER BY College DESC} & 100.0 \\
& What are the details of all organizations that are described as Sponsors and sort the results in ascending order? & \texttt{SELECT organisation\_details FROM Organisations AS T1 JOIN organisation\_Types AS T2 ON T1.organisation\_type  =  T2.organisation\_type WHERE T2.organisation\_type\_description  =  'Sponsor' ORDER BY organisation\_details} & 100.0 \\
& What are the nicknames of schools whose division is not 1? & \texttt{SELECT Nickname FROM school\_details WHERE Division != "Division 1"} & 1.0 \\
& What are the different names of the colleges involved in the tryout in alphabetical order? & \texttt{SELECT DISTINCT cName FROM tryout ORDER BY cName} & 0.6 \\
& What are the names of the members and branches at which they are registered sorted by year of registration? & \texttt{SELECT T3.name ,  T2.name FROM membership\_register\_branch AS T1 JOIN branch AS T2 ON T1.branch\_id  =  T2.branch\_id JOIN member AS T3 ON T1.member\_id  =  T3.member\_id ORDER BY T1.register\_year} & 100.0 \\
\midrule
\multirow{1}{=}{Random} & Count the number of students who have advisors. & \texttt{SELECT count(DISTINCT s\_id) FROM advisor} & 0.3 \\
& What is the number of aircraft? & \texttt{SELECT count(*) FROM aircraft} & 0.5 \\
& What is all the information about employees with D or S in their first name, ordered by salary descending? & \texttt{SELECT * FROM employees WHERE first\_name LIKE '\%D\%' OR first\_name LIKE '\%S\%' ORDER BY salary DESC} & 100.0 \\
& What is the id of the reviewer whose name includes the word "Mike"? & \texttt{SELECT rID FROM Reviewer WHERE name LIKE "\%Mike\%"} & 1.0 \\
& What is the average price for wines not produced in Sonoma county? & \texttt{SELECT avg(price) FROM wine WHERE Appelation NOT IN (SELECT T1.Appelation FROM APPELLATIONS AS T1 JOIN WINE AS T2 ON T1.Appelation  =  T2.Appelation WHERE T1.County  =  'Sonoma')} & 100.0 \\
& Find the average price of wines that are not produced from Sonoma county. & \texttt{SELECT avg(price) FROM wine WHERE Appelation NOT IN (SELECT T1.Appelation FROM APPELLATIONS AS T1 JOIN WINE AS T2 ON T1.Appelation  =  T2.Appelation WHERE T1.County  =  'Sonoma')} & 100.0 \\
& What are the names of all stations that have more than 10 bikes available and are not located in San Jose? & \texttt{SELECT T1.name FROM station AS T1 JOIN status AS T2 ON T1.id  =  T2.station\_id GROUP BY T2.station\_id HAVING avg(bikes\_available)  >  10 EXCEPT SELECT name FROM station WHERE city  =  "San Jose"} & 100.0 \\
& List the names of the customers who have once bought product "food". & \texttt{SELECT T1.customer\_name FROM customers AS T1 JOIN orders AS T2 JOIN order\_items AS T3 JOIN products AS T4 ON T1.customer\_id = T2.customer\_id AND T2.order\_id = T3.order\_id AND T3.product\_id = T4.product\_id WHERE T4.product\_name = "food" GROUP BY T1.customer\_id HAVING count(*)  >=  1} & 100.0 \\
\bottomrule
\caption{\textbf{Examples of 8 retrieved demonstrations for different selection methods}. The inference exemplar is the mapping \underline{\textit{What are the different conference names?}} $\mapsto$ \texttt{SELECT DISTINCT conference\_name FROM conference}. The QED score between the gold query and each retrieved query is shown as reference. Demonstrations retrieved with MARLO are metadata-agnostic. Although the question phrasing differs slightly and the domains are different, the SQL queries are structurally similar.}
\label{tab:retrieved-examples}
\end{longtable}
\end{scriptsize}
\pagebreak

\subsubsection{Reported Results for GPT 4 in Text-to-SQL}
\autoref{table:gpt4} lists execution accuracy of GPT 4 in Text-to-SQL reported by previous studies that implemented similarity-based demonstration selection methods. 
Note that only results on Spider development set are included in the table since most studies do not report results on the test set.
The DAIL-SQL results are shown as the SOTA benchmark. 
\begin{table*}
\caption{\textbf{Reference GPT 4 execution accuracy (\%) with ICL demonstration selection methods from literature}.
Results that are unavailable or impossible to reproduce are omitted using $\varnothing$. \textsuperscript{1}\textit{In cases where our evaluation conflicts with reported results, we reports ours instead.}}
\begin{center}
\begin{small}
\begin{adjustbox}{max width=\textwidth}
\begin{tabular}{clccc}
\toprule
\multirow{2}{*}{\textbf{LLM}} & \multirow{2}{*}{\textbf{Selection Method}} & \multirow{2}{*}{\textbf{\textnumero\text{ demos}.}} & \multicolumn{2}{c}{\textbf{Spider-dev}} \\ \cline{4-5}
 &  &  & \textbf{EX} & \textbf{EX\textsuperscript{\textdagger}} \\
\midrule
\multirow{11}{*}{GPT 4}
& Baseline \cite{pourreza2023dinsql} & 0  & 72.9\textsuperscript{1} & 77.8 \\ 
& Baseline \cite{gao2023texttosql} & 0  & 72.3 & $\varnothing$ \\ 
& Random \cite{chen2023teaching} & 32  & 73.2 & $\varnothing$ \\
& Random \cite{an-etal-2023-skill} & 4  & 76.1 & $\varnothing$ \\
& Random \cite{gao2023texttosql} & 5  & 79.5 & $\varnothing$ \\ 
& Random \cite{pourreza2023dinsql} & 6  & 76.8\textsuperscript{1} & 77.8 \\ 
& Question Similarity \cite{an-etal-2023-skill} & 4  & 76.7 & $\varnothing$ \\
& Question Similarity \cite{gao2023texttosql} & 5 & 79.9 & $\varnothing$ \\ 
& Masked Question Similarity \cite{gao2023texttosql} & 5 & 82.0 & $\varnothing$ \\ 
& Skills Similarity \cite{an-etal-2023-skill} & 4 & 82.7 & $\varnothing$ \\
& DAIL-SQL \cite{gao2023texttosql} & 9 & 83.1\textsuperscript{1} & 83.3\textsuperscript{1} \\
\bottomrule
\end{tabular}
\end{adjustbox}
\end{small}
\end{center}
\vskip -0.1in
\label{table:gpt4}
\end{table*}

\subsubsection{Breakdown of Results by Query Difficulty}
\begin{figure}[ht]
    \centering
    \includegraphics[width=0.8\linewidth]{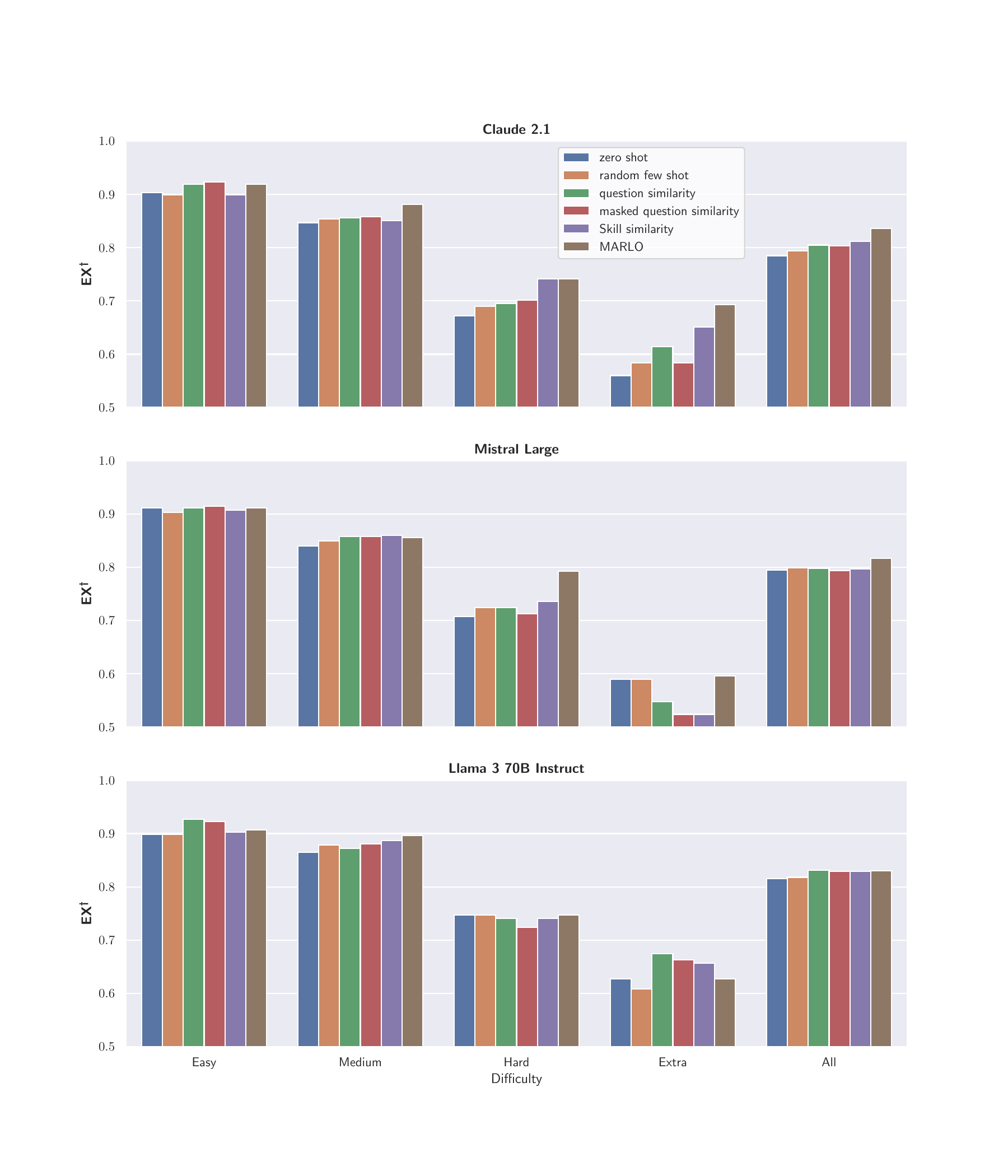}
    \caption{\textbf{Execution accuracy (\%) on Spider-dev by difficulty level.} With Claude 2.1 and Mistral Large, MARLO outperforms all other demonstration selection methods across all difficulty levels, particularly on more difficult questions, implying the examples to selects contributes to better question understanding in more complex settings.
    }
    \label{fig:breakdown}
\end{figure}

In \autoref{fig:breakdown} we compare execution accuracy ($\textbf{EX\textsuperscript{\textdagger}}$) across difficult levels on Spider-dev. For both Claude 2.1 and Mistral Large we see question understanding enabled by MARLO plays a pivotal role when generating the more difficult questions.

\subsection{Beyond Example Selection}
Current state-of-the-art Text-to-SQL systems\textsuperscript{\ref{footnote:sota}} comprise multiple stages such as schema linking and specialized decoding strategies \cite{wang2023selfconsistency} in addition to ICL with example selection.
Hence, we include two additional experiments to assess the effectiveness of MARLO with the most commonly used decoding strategy — self-consistency (SC) —
as well as an upper-bound (UB) estimate of the gains we can expect using an optimal strategy.
We sample 10 queries using MARLO with a temperature of 0.7 and report results in \autoref{table:marlo-sc}.
SC leads to minor performance gains— in-line with results reported in literature.
However, we observe a significant unrealized potential (+3-4\%pt.) comparing UB and SC across all models.
As it is not within the scope of this work, we encourage future work to explore preference optimization or voting strategies to boost LLM performance on this task.
\begin{figure}
\vspace{-0cm}
\caption{\textbf{Execution accuracy (\%) of MARLO sampling 10 predictions}. Self-consistency (SC) uses majority-voted predictions. Upper-bound (UB) uses the best predictions post-hoc. While SC leads to negligible gains, UB implies LLMs possess unrealized potential.}
\begin{small}
\begin{center}
\begin{adjustbox}{max width=\textwidth}
\begin{tabular}{clccccc}
\toprule
\multirow{2}{*}{\textbf{LLM}} & \textbf{Voting} & \multicolumn{2}{c}{\textbf{Spider-dev}} && \multicolumn{2}{c}{\textbf{Spider-test}}\\ \cline{3-4} \cline{6-7}
 & \textbf{Method} & \textbf{EX} & \textbf{EX\textsuperscript{\textdagger}} && \textbf{EX} & \textbf{EX\textsuperscript{\textdagger}} \\
\midrule
\multirow{3}{*}{\shortstack[c]{Claude\\2.1}} & - & 80.8 & 83.6 && 81.2 & 84.0 \\
& SC & 80.7 & 84.1 && 81.9 & 84.5 \\
& UB & 86.4 & 88.3 && 87.4 & 88.6 \\
\midrule
\multirow{3}{*}{\shortstack[c]{Mistral\\Large}} & - & 80.0 & 81.7 && 81.7 & 83.2 \\
& SC & 81.1 & 82.8 && 82.3 & 83.7 \\
& UB & 85.8 & 87.6 && 86.1 & 87.8 \\
\midrule
\multirow{3}{*}{\shortstack[c]{Llama 3\\70B\\Instruct}} & - & 82.2 & 83.0 && 83.2 & 83.8 \\
& SC & 83.3 & 84.0 && 83.7 & 84.5 \\
& UB & 88.0 & 88.7 && 87.1 & 87.9 \\
\bottomrule
\end{tabular}
\end{adjustbox}
\end{center}
\end{small}
\label{table:marlo-sc}
\vspace{-0cm}
\end{figure}

\subsection{Additional Training Cost \& Inference Latency}
\label{subsec:cost}
It is important to recognize that improved performance on the Text-to-SQL task using general-purpose LLMs often comes at the expense of additional training cost and inference latency.
The requirement to learn aligned embeddings by means of an additional fine-tuning process sets this approach apart from most other example selection baselines.
However, compared to the approaches that rely on generic embeddings, MARLO is able to achieve considerable performance gains at the expense of minimal training cost overhead by fine-tuning a relatively small encoder.
Since the embedding dimension remains the same across all selection methods, MARLO does not incur additional inference latency compared to other example selection methods that rely on vector search.
However, compared to other approaches, such as Skill-KNN or DAIL-SQL\textsuperscript{\ref{footnote:sota}} that require preliminary or additional LLM calls during inference, MARLO is able to achieve comparable performance with considerably lower inference latency.

\newpage
\begin{longlisting}
\caption{\textbf{Example SQL generation prompt with a single in-context example}. The language model generates the predicted query by autoregressively completing the prompt until the close SQL XML tag (\texttt{</sql>}) is generated. For zero-shot inference, the example blocks and introductory heading are simply omitted.}
\begin{small}
\begin{minted}[mathescape,
linenos=False,
bgcolor=LightGray,
frame=lines,
breaklines]{yaml}
Human: Paying careful attention to the table and column names in the given metadata, provide a correct SQLite query to answer the given question. Enclose your query in '<sql></sql>' XML tags.

Here are some examples:

<example>
Metadata:
<metadata>
CREATE TABLE bank (
branch_ID int PRIMARY KEY,
bname varchar(20),
no_of_customers int,
city varchar(10),
state varchar(20))
CREATE TABLE customer (
cust_ID varchar(3) PRIMARY KEY,
cust_name varchar(20),
acc_type char(1),
acc_bal int,
no_of_loans int,
credit_score int,
branch_ID int,
state varchar(20),
FOREIGN KEY(branch_ID) REFERENCES bank(branch_ID))
CREATE TABLE loan (
loan_ID varchar(3) PRIMARY KEY,
loan_type varchar(15),
cust_ID varchar(3),
branch_ID varchar(3),
amount int,
FOREIGN KEY(branch_ID) REFERENCES bank(branch_ID),
FOREIGN KEY(Cust_ID) REFERENCES customer(Cust_ID))

1 sample row from "bank" table:
"""
bname: downtown
branch_ID: 2
city: Salt Lake City
no_of_customers: 123
state: Utah
"""
1 sample row from "customer" table:
"""
acc_bal: 2000
acc_type: saving
branch_ID: 2
credit_score: 30
cust_ID: '1'
cust_name: Mary
no_of_loans: 2
state: Utah
"""
1 sample row from "loan" table:
"""
amount: 2050
branch_ID: '1'
cust_ID: '1'
loan_ID: '1'
loan_type: Mortgages
"""
</metadata>
Question: Count the number of bank branches.
SQL Query: <sql>SELECT count(*) FROM bank</sql>
</example>

Metadata:
<metadata>
CREATE TABLE Ref_Template_Types (
Template_Type_Code CHAR(15) NOT NULL,
Template_Type_Description VARCHAR(255) NOT NULL,
PRIMARY KEY (Template_Type_Code)
)
CREATE TABLE Templates (
Template_ID INTEGER NOT NULL,
Version_Number INTEGER NOT NULL,
Template_Type_Code CHAR(15) NOT NULL,
Date_Effective_From DATETIME,
Date_Effective_To DATETIME,
Template_Details VARCHAR(255) NOT NULL,
PRIMARY KEY (Template_ID),
FOREIGN KEY (Template_Type_Code) REFERENCES Ref_Template_Types (Template_Type_Code)
)
CREATE TABLE Documents (
Document_ID INTEGER NOT NULL,
Template_ID INTEGER,
Document_Name VARCHAR(255),
Document_Description VARCHAR(255),
Other_Details VARCHAR(255),
PRIMARY KEY (Document_ID),
FOREIGN KEY (Template_ID) REFERENCES Templates (Template_ID)
)
CREATE TABLE Paragraphs (
Paragraph_ID INTEGER NOT NULL,
Document_ID INTEGER NOT NULL,
Paragraph_Text VARCHAR(255),
Other_Details VARCHAR(255),
PRIMARY KEY (Paragraph_ID),
FOREIGN KEY (Document_ID) REFERENCES Documents (Document_ID)
)

1 sample row from "Ref_Template_Types" table:
"""
Template_Type_Code: CV
Template_Type_Description: CV
"""
1 sample row from "Templates" table:
"""
Date_Effective_From: '1996-02-04 11:27:24'
Date_Effective_To: '1995-09-19 22:27:48'
Template_Details: ''
Template_ID: 11
Template_Type_Code: BK
Version_Number: 6
"""
1 sample row from "Documents" table:
"""
Document_Description: z
Document_ID: 33930
Document_Name: How Google people work
Other_Details: null
Template_ID: 1
"""
1 sample row from "Paragraphs" table:
"""
Document_ID: 651512
Other_Details: null
Paragraph_ID: 243399026
Paragraph_Text: Indonesia
"""
</metadata>
Question: How many paragraphs in total?

Assistant: SQL Query: <sql>
\end{minted}
\label{lst:example-prompt}
\end{small}
\end{longlisting}

\end{document}